%% file: camera_ready.tex
\documentclass{article}
\PassOptionsToPackage{numbers,sort&compress}{natbib}
\usepackage[final]{neurips_2026} 
\makeatletter
\renewcommand{\@noticestring}{Workshop on Who Verifies the Agents? Toward Reliable Agent Development, NeurIPS 2026.}
\makeatother
\input{preamble}
\namedbuildtrue 

\newcommand{\repourl}{\footnote{\url{https://github.com/sumanyumuku98/MemMux}}}

\author{%
  Sumanyu Muku \\
  Amira Learning \\
  \texttt{mukusumanyu@gmail.com} \\
}

\begin{document}
\maketitle
\input{body}

\end{document}

%% file: preamble.tex
\usepackage[T1]{fontenc}
\usepackage{lmodern}
\usepackage{microtype}
\usepackage{hyperref}
\usepackage{url}
\usepackage{booktabs}
\usepackage{capt-of} 
\usepackage{amsmath,amssymb}
\usepackage{graphicx}
\usepackage{xcolor}
\usepackage{textcomp}
\usepackage{upquote} 
\usepackage{listings}
\usepackage{pgfplots}
\pgfplotsset{compat=1.17}
\usetikzlibrary{positioning,arrows.meta,shapes.geometric,fit,backgrounds,calc}

\newcommand{\cmark}{{\color{syscol}\checkmark}}
\newcommand{\xmark}{{\color{inkcol!40}\ensuremath{\times}}}

\newif\ifnamedbuild
\namedbuildfalse

\definecolor{rawcol}{HTML}{DC2626}   
\definecolor{tmuxcol}{HTML}{D97706}  
\definecolor{herdrcol}{HTML}{0D9488} 
\definecolor{syscol}{HTML}{2563EB}   
\definecolor{accentcol}{HTML}{6D28D9}
\definecolor{inkcol}{HTML}{1F2937}   
\definecolor{panelcol}{HTML}{F3F4F6} 

\hypersetup{colorlinks=true, linkcolor=syscol, citecolor=accentcol,
  urlcolor=syscol, filecolor=syscol}

\pgfplotsset{
  memmux/.style={
    width=\linewidth, height=5.2cm,
    tick align=outside, tick pos=left,
    axis line style={inkcol!55}, axis lines=left,
    grid=major, grid style={inkcol!10, line width=0.4pt},
    label style={font=\small}, tick label style={font=\footnotesize},
    legend style={font=\footnotesize, draw=inkcol!20, rounded corners=2pt,
      fill=white, fill opacity=0.92, text opacity=1, inner sep=3pt},
    legend cell align=left,
    every axis plot/.append style={line width=1.2pt},
    mark size=2.4pt,
    clip=false,
  },
}

\lstdefinestyle{shell}{
  basicstyle=\ttfamily\footnotesize\color{inkcol},
  backgroundcolor=\color{panelcol},
  frame=leftline, framerule=1.2pt, rulecolor=\color{syscol!55},
  xleftmargin=8pt, framexleftmargin=8pt, aboveskip=6pt, belowskip=2pt,
  breaklines=true, breakatwhitespace=true, columns=fullflexible,
  keepspaces=true, showstringspaces=false, upquote=true,
}

\tikzset{
  comp/.style={rounded corners=2.5pt, draw=syscol!45, fill=white,
    inner sep=4pt, font=\scriptsize, align=center, minimum height=7mm},
  taskbox/.style={rounded corners=4pt, draw=inkcol!25, fill=inkcol!3,
    inner sep=5pt},
  proc/.style={rounded corners=2pt, draw=inkcol!35, fill=white,
    font=\scriptsize, inner sep=2.5pt, align=center, minimum width=13mm},
  root/.style={proc, draw=syscol!60, fill=syscol!8, font=\scriptsize\bfseries},
  escaped/.style={proc, draw=rawcol!70, fill=rawcol!7, font=\scriptsize},
  client/.style={rounded corners=3pt, draw=inkcol!45, fill=panelcol,
    inner sep=5pt, font=\scriptsize, align=center},
  flow/.style={-{Stealth[length=5pt]}, inkcol!75, line width=0.9pt},
  biflow/.style={{Stealth[length=5pt]}-{Stealth[length=5pt]}, inkcol!75, line width=0.9pt},
  scan/.style={-{Stealth[length=5pt]}, dashed, syscol!75, line width=0.9pt},
  esc/.style={-{Stealth[length=5pt]}, dashed, rawcol!75, line width=0.9pt},
}

\newcommand{\sys}{\textsc{MemMux}}
\newcommand{\code}[1]{\texttt{#1}}
\title{\sys{}: Runtime Verification and Honest Resource Attribution\\
for Fleets of Parallel Coding Agents}

%% file: body.tex

\begin{abstract}
Developers increasingly run not one coding agent but a fleet of them side by side on one
workstation. The tools they reach for, terminal multiplexers like \code{tmux} and a new generation of
agent managers, were built to arrange windows, not to govern memory. When ten agents each spawn
language servers, test runners, and browsers, no standard tool in that stack can say how much memory
belongs to which agent, confirm that a terminated agent's descendants are gone, notice a child that
has escaped its agent, or keep the machine off the swap cliff when an out-of-memory (OOM) kill would silently discard
uncommitted work. We treat these as runtime-verification problems. An agent-hosting substrate should
continuously emit observable signals that an operator or auditor can check while agents run. We
present \sys{}, an unprivileged local runtime---no kernel control groups---that turns resource
governance into a set of checkable signals
(per-agent attribution, complete reclamation, escaped-process visibility, bounded footprint under
overcommit, and monitoring overhead) and a claims-disciplined benchmark that measures each one against
\code{tmux} and a raw-process baseline on identical workloads, with a purpose-built agent multiplexer
where it could be driven.
Under a binding memory budget on a Linux host, \sys{} keeps the fleet under budget (7500\,MiB) with zero
swap by admitting a subset and reclaiming under pressure. The ungoverned tools instead run every agent,
pin the machine at its RAM ceiling ($2\times$ over budget), and spill about 2\,GiB into swap. \sys{}
reclaims 100\% of a terminated agent's process subtree where the raw baseline strands half of it (ten
orphaned processes, 303\,MiB; \code{tmux} matches us here, the raw baseline does not), and it is the
only launcher we compare against that surfaces an escaped child, detecting all
10 of 10 injected escapes. The cost is real and we report it: the always-on 1\,Hz monitoring loop
runs near 0.6\% CPU for one agent but reaches 2.7\% at ten, above our 2\% target. Running the same
harness on real Claude Code sessions shows that 100\% attribution and low overhead carry over to live,
multi-process agent trees. We release the engine, the benchmark, and a one-command reproducer.
\end{abstract}

\section{Introduction}

A year ago, running an autonomous coding agent meant babysitting a single session. Today it is
normal to fan a task out across several agents at once---one refactoring, one writing tests, one
chasing a flaky build---and to watch them in a grid of panes. The agents themselves have improved
quickly~\citep{yao2023react,shinn2023reflexion,jimenez2024swebench,yang2024sweagent}; the substrate
they run on has not. Developers arrange these fleets with terminal multiplexers such as
\code{tmux}~\citep{tmux} or with purpose-built agent managers, and both were designed to place
windows on a screen, not to answer questions about memory.

Those questions matter because a coding agent is not a single process. It is a tree: the model's
CLI, a shell, language servers, test and build workers, sometimes a headless browser. A handful of
such trees can exhaust a laptop's RAM. When they do, the operating system's response---the Linux OOM
killer, or swap thrashing---is blunt and invisible: a process is killed mid-write, or the whole
machine grinds, and the agent's uncommitted edits are simply gone. Nothing in the stack was
accountable for that memory in the first place. \code{tmux} does not know that pane 7's language
server belongs to the same logical agent as pane 3's test runner; it cannot tell you that a
double-forked child has escaped supervision; and it has no budget it is trying to keep.

We think the right lens for this is \emph{runtime verification}~\citep{bartocci2018introrv}: rather
than prove properties of the substrate ahead of time, instrument it so the properties it claims can
be \emph{checked while it runs}, and surface any violation rather than hide it. Cluster schedulers
made an analogous move a decade ago---Borg combined admission control, over-commitment, and
process-level isolation, and treated monitoring as a first-class output~\citep{verma2015borg}. We
bring that stance down to a single developer machine and to the specific failure modes of agent
fleets.

This reframes an infrastructure problem as a verification one, on an axis the agent-evaluation
literature largely leaves open. Most work on verifying agents asks whether an agent's \emph{outputs}
are trustworthy: did it resolve the issue, is the reward genuine, is the behaviour aligned. We ask a
different question, one that matters more once agents run in parallel and unattended. Does the
\emph{substrate that hosts them} uphold operational guarantees an operator can check as the fleet
runs: accurate accounting, complete reclamation, and no silently lost work? These are always-on
signals over the live execution, not one-shot scores. A fleet whose host cannot answer them is
unverified, however capable each agent is. This is a precondition for verifying agents rather
than a detour around it: an agent's observed behaviour is attributable and reproducible only if the
host did not silently swap it, starve it, or OOM-kill it mid-task. A trace collected on an unverified
substrate cannot separate a model that failed from a machine that did.

This paper contributes three things. First (Section~\ref{sec:signals}), \textbf{five verification
signals for agent fleets}: per-agent attribution, complete reclamation on teardown, escaped-process
visibility, bounded footprint (with preserved work) under overcommit, and monitoring overhead. Four
are runtime properties; the fifth is the monitor's own cost. Each has a stated threshold. Second
(Section~\ref{sec:design}), \textbf{\sys{}}, a local runtime that observes the signals and enforces
the governance properties. Third (also Section~\ref{sec:method}), \textbf{a claims-disciplined benchmark}
that drives \code{tmux}, an agent multiplexer, and a raw baseline through \emph{identical}
deterministic workloads, tags every process as agent or manager overhead, and never reports a number
it did not measure. A competitor that cannot be driven fairly is recorded as skipped, not estimated.
We evaluate across two Linux hosts; on the smaller one a binding-budget experiment separates \sys{} from the ungoverned tools by
keeping a twelve-agent stress workload under budget and off swap (Section~\ref{sec:eval});
Section~\ref{sec:limits} states the limits: mostly synthetic workloads, fleets no larger than twelve,
and an overhead target not yet met.

\section{Background and Related Work}

\paragraph{Language-model agents.} Interleaving reasoning with tool use~\citep{yao2023react} and
adding self-reflection~\citep{shinn2023reflexion} turned language models into agents that take long
sequences of actions. Coding is the canonical proving ground: SWE-bench frames the task as resolving
real GitHub issues against a repository~\citep{jimenez2024swebench}, and SWE-agent shows a
well-designed agent-computer interface matters as much as the model~\citep{yang2024sweagent}.
Multi-agent settings, first studied for open-ended behaviour~\citep{park2023generative}, are now a
practical reality on developer machines. This work is orthogonal to the agents: we govern whatever
they spawn.

\paragraph{Terminal multiplexers.} \code{tmux}~\citep{tmux} is the default way developers run many
long-lived sessions at once. It is excellent at persistent panes and detach/reattach, and it is the
baseline any ``run many agents'' tool must beat. But a pane is just a process group under a server,
with no notion of a logical agent, a memory budget, or a process that has wandered out of its
subtree.

\paragraph{Resource management.} Cluster managers solved fleet-scale governance long ago: Borg with
admission control and over-commitment~\citep{verma2015borg}, Mesos with two-level
scheduling~\citep{hindman2011mesos}, and Kubernetes as the open-source lineage~\citep{kubernetes}.
The enforcement primitives they use---Linux control groups~\citep{cgroupv2} and the OOM
killer~\citep{linuxoom}---live in every modern kernel. \sys{} borrows the ideas (reserve, admit,
reclaim before you thrash) but targets a single un-orchestrated laptop, and it deliberately avoids
hard cgroup containment in favor of signal-based supervision. This is an explicit trade-off: cgroups
give hard, kernel-enforced limits but need privilege and per-machine setup, whereas signal-based
supervision runs unprivileged and portably (macOS included) at the cost of best-effort rather than
hard containment---a child can escape between samples, and one that escapes before it is ever
observed is not caught at all (H3; the five signals are defined in \S\ref{sec:signals}). We take that trade knowingly. A kernel cgroup bounds a footprint
harder than 1\,Hz supervision can, and where privilege and delegation are available it is the
stronger enforcement mechanism. What it does not do is decline to start work that will not fit, or
preserve uncommitted work before destroying a process; and it is unavailable unprivileged and absent
on macOS. A quantified head-to-head against a \code{systemd}-scope baseline under \code{MemoryMax}
would separate what \sys{} contributes from what the kernel already provides, and is the clearest
next experiment.

\paragraph{Memory accounting.} Summing resident set size (RSS) over a process tree double-counts
shared pages. The proportional set size (PSS) exposed through \code{/proc/<pid>/smaps\_rollup}
divides each shared page by its sharers and is the fair per-process figure~\citep{procfs}; macOS
offers \code{phys\_footprint} as its analogue. \sys{} attributes by PSS (RSS fallback) so a task's
number reflects memory it is responsible for.

\paragraph{Runtime verification.} Rather than verify a system statically, runtime verification
instruments it and checks a specification against the execution trace, reacting to
violations~\citep{bartocci2018introrv}. Our first four signals are such monitors: cheap, always-on
checks with explicit thresholds whose failures are events, not silence; the fifth quantifies their
runtime cost.

\section{The Design of \sys{}}
\label{sec:design}

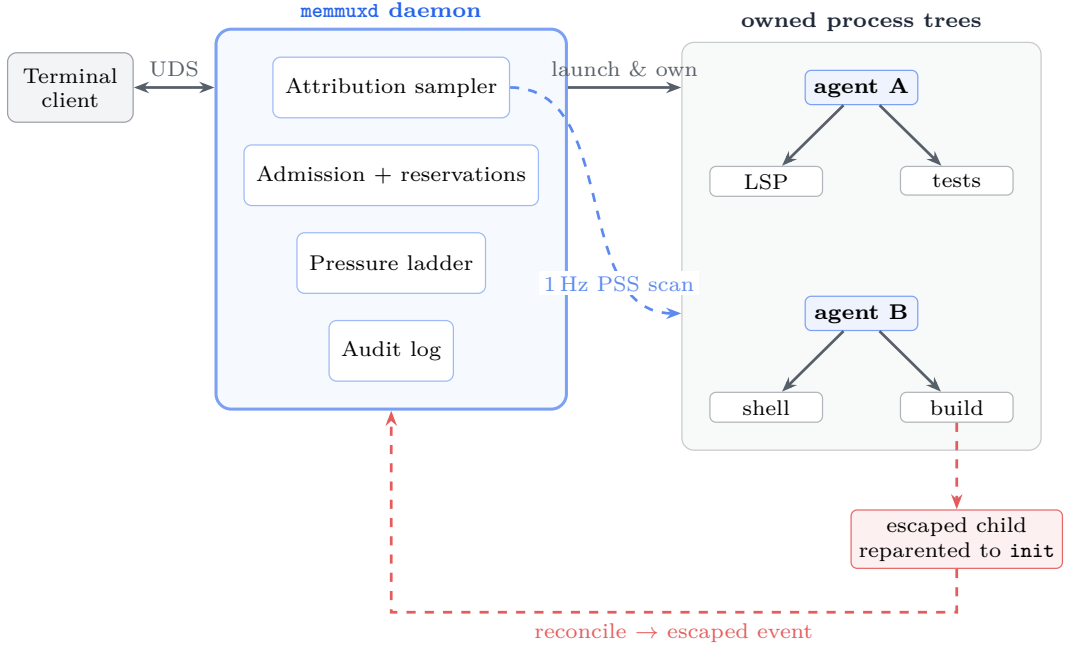
\begin{figure}[t]
\centering
\resizebox{\linewidth}{!}{%
\begin{tikzpicture}[node distance=3mm and 16mm]
  \node[client] (client) {Terminal\\client};
  \node[comp, right=of client] (attr)   {Attribution sampler};
  \node[comp, below=of attr]   (adm)    {Admission $+$ reservations};
  \node[comp, below=of adm]    (ladder) {Pressure ladder};
  \node[comp, below=of ladder] (audit)  {Audit log};
  \begin{scope}[on background layer]
    \node[rounded corners=5pt, draw=syscol!60, line width=1pt, fill=syscol!4,
          fit=(attr)(adm)(ladder)(audit), inner sep=9pt] (daemon) {};
  \end{scope}
  \node[font=\scriptsize\bfseries\color{syscol}, anchor=south] at (daemon.north)
        {\code{memmuxd} daemon};
  \node[root, right=34mm of attr] (rA) {agent A};
  \node[proc, below=7mm of rA, xshift=-11mm] (a1) {LSP};
  \node[proc, below=7mm of rA, xshift=11mm]  (a2) {tests};
  \node[root, below=22mm of rA] (rB) {agent B};
  \node[proc, below=7mm of rB, xshift=-11mm] (b1) {shell};
  \node[proc, below=7mm of rB, xshift=11mm]  (b2) {build};
  \begin{scope}[on background layer]
    \node[taskbox, fit=(rA)(a1)(a2)(rB)(b1)(b2), inner sep=9pt] (trees) {};
  \end{scope}
  \node[font=\scriptsize\bfseries\color{inkcol}, anchor=south] at (trees.north)
        {owned process trees};
  \node[escaped, below=10mm of b2] (esc) {escaped child\\reparented to \code{init}};
  \draw[flow] (rA) -- (a1); \draw[flow] (rA) -- (a2);
  \draw[flow] (rB) -- (b1); \draw[flow] (rB) -- (b2);
  \draw[esc]  (b2) -- (esc);
  \draw[biflow] (client.east) -- node[above,font=\scriptsize]{UDS} (daemon.west|-client);
  \draw[flow] (daemon.east|-rA) -- node[above,font=\scriptsize]{launch \& own} (trees.west|-rA);
  \draw[scan] (attr.east) to[out=0,in=180]
        node[below,font=\scriptsize,pos=0.72,fill=white,inner sep=1.5pt]{1\,Hz PSS scan} (trees.west|-rB);
  \draw[esc] (esc.south) -- ++(0,-5mm) -|
        node[below,font=\scriptsize,pos=0.25]{reconcile $\to$ escaped event} (daemon.south);
\end{tikzpicture}%
}
\caption{\sys{} architecture. A terminal client drives the \code{memmuxd} daemon over a Unix
domain socket. The daemon launches and \emph{owns} each agent's process tree; a 1\,Hz sampler sums
proportional memory (PSS) across every owned subtree (H1), reconciliation flags any child that
double-forks out to \code{init} (H3), and admission, the pressure ladder, and an audit log turn the
budget into checkable, logged decisions (H4).}
\label{fig:arch}
\end{figure}

\sys{} is a Rust workspace: a daemon (\code{memmuxd}) that owns state and does the governing, and a
terminal client that attaches over a Unix domain socket (UDS). A \emph{task} is a durable, logical agent;
a \emph{runtime} is one physical launch of a provider inside a pseudo-terminal. The design is
organized around making each signal checkable (Figure~\ref{fig:arch}). Enforcement throughout is
signal-based and unprivileged: \sys{} uses no kernel control groups, so every budget below is held by
admission and reclamation decisions rather than by kernel containment.

\paragraph{Attribution.} Once per second the daemon snapshots the host process tree and, for each
running task's provider root, sums the PSS of the root and all descendants. That per-task figure
feeds three consumers: the budget, the client's task view, and the escaped-process check (which uses
the pids recorded under each root). Attribution is scoped to processes \sys{} launched, so unrelated
host activity never pollutes a task's number.

\paragraph{Recursive reclamation.} When a task is terminated (by the operator or because its provider
exited), \sys{} reaps the \emph{entire} owned subtree. Order matters: it signals the subtree while the
root is still alive, because once the root dies its children reparent to \code{init} and the tree that
identifies them is lost. For a provider that already exited, \sys{} reaps from the pids it recorded
while the task ran. The target is at least 99.5\% of owned descendants gone within ten seconds.

\paragraph{Escaped-process reconciliation.} A child that double-forks and reparents to \code{init}
has escaped its agent. This is the classic way a leak hides. Each sample, \sys{} reconciles the pids
seen under a task against the live tree; a once-owned pid now living outside every task subtree is
flagged \emph{escaped} and surfaced as an event. Reconciliation is scoped to the managed set, so the signal
is a real ``we lost track of a process we started,'' not host noise.

\paragraph{Admission and the pressure ladder.} The daemon holds a memory budget (physical RAM minus
reserves for the OS, interactive apps, and a swap-safety margin). Starting a task reserves its
predicted peak; if the reservation would exceed the budget the task stays queued with a visible
reason rather than over-subscribing the machine. When live usage climbs anyway, a graduated
\emph{pressure ladder} responds before the machine thrashes: reclaim escaped children, hibernate the
lowest-priority idle task, recycle the largest provider, and, at the top of the ladder, terminate the
lowest-priority tree, but only after persisting a checkpoint of the victim's git state. Preserving
work always precedes destroying a process. Every governance action writes an audit decision with its
reason and evidence, so the system is auditable, not just automatic. These decisions are the only
thing holding the budget---there is no cgroup beneath it---which is what makes bounded footprint a
measured property here rather than a kernel-guaranteed one.

\section{Verification Signals and Methodology}
\label{sec:signals}
\label{sec:method}

We frame governance as runtime monitors over the fleet's execution. The observed \emph{trace} is the
sequence of once-per-second process-tree samples and lifecycle events; each signal is a
\emph{monitor} over that trace with an explicit threshold (Table~\ref{tab:signals}), and a breach is
surfaced as a violation event rather than silently tolerated. Four are properties of the running system
(H1--H4); H5 is the \emph{cost} of monitoring, not an invariant, and H4 bundles three
sub-properties---admission, bounded footprint, and preservation of uncommitted work---which we report
separately below as \textbf{H4a} (admission and bounded footprint) and \textbf{H4b} (preservation of
uncommitted work ahead of a destructive termination); the generated run reports use the same labels. \sys{} both \emph{observes} these signals and \emph{enforces} them (admission,
reclamation); the baselines can only be observed. These are empirical monitors with measured pass/fail
behaviour, not machine-checked proofs, so we claim the signals are checkable and checked, not that the
engine is sound and complete in the formal sense. A benchmark harness drives every launcher through
identical, deterministic workloads and evaluates the monitors from one external process sampler---an
\emph{oracle} using the daemon's own accounting code, pointed at whatever root pids a launcher reports,
so measurement is product-agnostic.

\begin{table}[t]
\centering\small
\begin{tabular}{@{}llp{7.2cm}@{}}
\toprule
& \textbf{Signal} & \textbf{Monitor / threshold} \\
\midrule
H1 & Attribution        & native signal maps $\geq 95\%$ of a launched tree's proportional memory to its owning task \\
H2 & Cleanup            & $\geq 99.5\%$ of an owned subtree gone $\leq 10$\,s after the launcher's normal teardown \\
H3 & Escaped visibility & every previously observed process that escapes and remains live through the next sample is surfaced as an event \\
H4 & Bounded / no-loss  & peak footprint $\leq 1.1\times$ budget under overcommit; uncommitted work checkpointed before destructive termination \\
H5 & Overhead           & continuous monitoring costs $\leq 2\%$ CPU at $N{=}10$ \\
\bottomrule
\end{tabular}
\caption{The five runtime-verification signals and their pass thresholds.}
\label{tab:signals}
\end{table}

\paragraph{Launchers.} Each launcher starts $N$ identical agents and reports the pids that constitute
the fleet: a \emph{raw} baseline (spawn the stub directly), \code{tmux} (one pane per agent on an
isolated server), and \sys{} (create $N$ tasks through the real daemon). A fourth, the \code{herdr}
agent multiplexer, is driven through its socket API where installed; on our stock Linux reference host
it was absent and is reported as skipped rather than guessed, as are \code{dmux} (which requires a
controlling terminal and cannot be driven headlessly), \code{cmux}, and \code{agentmux}.

\paragraph{Workload.} A deterministic stub agent replays a recording of allocate / spawn-child / emit
/ sleep steps, touching every page it allocates so its resident memory is real. We use a stub rather
than a live model on purpose: runs are then reproducible and identical across launchers, at the cost
of realism (Section~\ref{sec:limits}). A \emph{hold} scenario keeps $N$ agents, each with a live
child, resident together for the measurement window. Without it, short workloads finish before a
fleet can be observed, an error we hit early and corrected.

\paragraph{Claims discipline.} The harness records each launcher's version, the measurement date, and
the host; it tags every sampled process as agent or manager overhead so the two are reported
separately; and it emits a number only for a monitor it actually measured. A launcher that cannot be
driven contributes ``unsupported,'' never a fabricated zero. These rules are part of the
contribution: there is no accepted methodology for comparing agent-hosting tools, and an unfair
comparison is worse than none. One point needs care. Because the oracle can attribute \emph{any}
launcher's tree after the fact, per-agent attribution is \emph{measurable} for all of them, and the
real-agent run reports 100\% for every launcher (Table~\ref{tab:realagent}). H1 as a signal asks the
stricter question of whether a launcher exposes that attribution \emph{natively and continuously} to
an operator. Only \sys{} does; \code{tmux} and the raw baseline surface no such figure themselves.
Table~\ref{tab:matrix} marks that native capability, not the oracle's coverage.

\section{Evaluation}
\label{sec:eval}

\paragraph{Setup.} Signals H1--H3 and H5 run on an AWS \code{m7i.2xlarge} reference host (8 vCPUs,
Intel Xeon Platinum 8488C, $\approx$31\,GiB RAM, Ubuntu 24.04), sweeping $N \in \{1,5,10\}$ over two
trials. The bounded-footprint claim (H4) needs \emph{real} memory pressure, so we run it on a
deliberately smaller 15.3\,GiB host (\code{m7i.xlarge}, 4 vCPUs, 12\,GiB swap) with a calibrated
synthetic stress footprint: each agent allocates 1400\,MiB, matching the reservation prior of
\sys{}'s default \emph{Standard} task class, so admission is well-calibrated (reservation
$\approx$ actual) rather than over-reserving; each resulting process tree holds $\sim$1.5\,GiB, and
$N{=}12$ of them ($\sim$18\,GiB) genuinely overcommit the machine. The H1--H3/H5 sweeps use two
trials as reproducibility checks and report their means without population-level confidence claims.
The H4 run uses five trials; its archived report gives mean\,$\pm$\,95\% confidence intervals for footprint, while
Table~\ref{tab:h4} rounds the means and reports observed peak swap. Signals H1--H3 and H5 were measured
on \sys{} 0.8.0, H4 on 0.9.0 (the release that added the configurable stress footprint), and the
H4b checkpoint demonstration on 0.9.1; these differ in benchmark tooling and in the checkpoint's
target directory rather than in the governance paths under test, but no single build has been
measured across all five signals in one sweep. A macOS
cross-check (Apple M4 Max, 36\,GiB) that also exercised \code{herdr} agreed qualitatively: like the
other ungoverned managers, \code{herdr} reaped its subtrees on teardown but saw no escapes and grew
with the fleet. Table~\ref{tab:matrix} summarizes the head-to-head; the subsections that follow are
the evidence behind each cell.

\begin{table}[t]
\centering\small
\setlength{\tabcolsep}{5pt}
\begin{tabular}{@{}lcccrrc@{}}
\toprule
Launcher & Attrib. & Cleanup & Escaped & Peak footprint (MiB) & Swap (MiB) & Ckpt. \\
\midrule
raw baseline    & \xmark & 50\%          & \xmark & 15075 & 1882 & \xmark \\
\code{tmux} 3.4 & \xmark & \textbf{100\%}& \xmark & 15347 & 2165 & \xmark \\
\sys{}$^\ddagger$    & \cmark & \textbf{100\%}& \cmark~10/10 & \textbf{7479} & \textbf{0} & \cmark$^\dagger$ \\
\bottomrule
\end{tabular}
\caption{Head-to-head on the signals (Linux; see \S\ref{sec:eval} for the per-signal hosts). \sys{} is the only launcher that
natively attributes per-agent memory (H1), surfaces escaped processes (H3), holds the fleet
\emph{under} a binding memory budget with zero swap (H4), and can checkpoint dirty work before
destructive termination. It \emph{matches} \code{tmux} on teardown cleanup (H2). Footprint/swap are peaks under a
binding 7500\,MiB budget on a 15.3\,GiB host ($N{=}12$); the ungoverned tools overcommit to the RAM
ceiling and spill $\sim$2\,GiB into swap. Attribution is the \emph{native} capability, not the
oracle's coverage (\S\ref{sec:method}); \sys{} also pays a monitoring cost the baselines do not (H5,
\S\ref{sec:h1h5}). $^\dagger$Checkpoint-before-destructive-termination is exercised end-to-end on a
single task (\S\ref{sec:h4}); automated resume from that checkpoint is not.
$^\ddagger$H1--H3/H5 use \sys{} 0.8.0; H4 and the live forced-kill run use 0.9.0; the H4b
demonstration uses 0.9.1. Each cell
is the headline result for that signal.}
\label{tab:matrix}
\end{table}

\subsection{H4a and H4b: bounded footprint and preserved work under overcommit}
\label{sec:h4}

We make the budget \emph{binding} using the calibrated stress workload of \S\ref{sec:eval}:
$N{=}12$ agent trees ($\sim$18\,GiB) against a 15.3\,GiB host, with \sys{} given a 7500\,MiB
budget---roughly half what the fleet wants. Table~\ref{tab:h4} reports peak total footprint and peak
swap over five trials. The ungoverned launchers run all twelve agents: their resident set pins the
machine at its $\sim$15\,GiB RAM ceiling---$2\times$ over the budget---and spills 1.8--2.1\,GiB into
swap. \sys{} instead governs, holding peak footprint at 7479\,MiB (\emph{under} budget), settling to
4512\,MiB steady, with \textbf{zero} swap. The margin is narrow---21\,MiB against a 7500\,MiB
budget---and it is a \emph{sampled} peak: a 1\,Hz observer can miss a brief excursion between two
ticks, so the figure bounds what was observed rather than what was reached. A kernel-provided
high-water mark would settle that directly and is the obvious next measurement. The zero-swap result
is less exposed to the same objection, since swapped pages persist across ticks rather than appearing
only between them. The run recorded seven deferred starts and twelve reclamation events in total; these are event
counts rather than unique tasks.

\begin{table}[t]
\centering\small
\begin{tabular}{@{}lrrr@{}}
\toprule
Launcher & Peak (MiB) & Steady (MiB) & Peak swap (MiB) \\
\midrule
raw baseline & 15075 & 15069 & 1882 \\
\code{tmux} 3.4     & 15347 & 15069 & 2165 \\
\sys{} 0.9.0 & \textbf{7479} & \textbf{4512} & \textbf{0} \\
\bottomrule
\end{tabular}
\caption{Binding-budget \emph{overcommit} ($N{=}12$ agent trees, $\sim$1.5\,GiB each, 15.3\,GiB host, five
trials). \sys{}'s 7500\,MiB budget is half the $\sim$15\,GiB the ungoverned fleet uses; \sys{} holds
peak footprint under budget with zero swap, while raw and \code{tmux} overcommit the machine
$2\times$ and spill $\sim$2\,GiB into swap.}
\label{tab:h4}
\end{table}

The trade-off is throughput. \sys{} admits at most about five agents and can settle lower as pressure
acts, so enforcing the budget means not running every agent at once. That is what a budget is for: the
ungoverned tools run all twelve only by driving the machine into swap, where an OOM kill can strike.
We do not measure the resulting completion time. Admission converts a memory overrun into queueing
delay, and quantifying that exchange---makespan under governance against the wall-clock cost of an
OOM kill and the work it destroys---is the comparison a user deciding whether to adopt governance
would want, and we do not yet provide it.
We do, however, address one half of that cost directly: the work an OOM kill destroys.

\textbf{H4b: preserved work.} Before the pressure ladder's \emph{emergency}
stage destructively terminates a victim, it first captures a durable checkpoint (git HEAD, dirty manifest, and patch
hash). The twelve-agent run above never reached emergency
termination---its reclamations were hibernate/recycle actions---so we exercised that path directly.
Driving a single task with a dirty worktree past a binding budget until the ladder reaches
\emph{emergency} (\sys{} 0.9.1, one trial, macOS), the daemon writes a checkpoint carrying the
victim's real patch hash with its dirty flag set, and only then terminates the tree, reclaiming
100\% of it. The recorded hash matches the one the independent worktree-retention path computes for
the same tree, so the two agree on what was preserved. Recovery today means that retained worktree
together with its checkpoint: automated \emph{resume} is wired for hibernated tasks, not for
emergency-terminated ones, so this is preservation rather than one-command restoration. Ungoverned
launchers have no checkpoint mechanism at all.

\subsection{H2: complete cleanup on teardown}

Each launcher starts a fleet whose agents each hold a memory-bearing child, then is torn down exactly
as it normally would be (raw: kill the roots; \code{tmux}: \code{kill-server}; \sys{}: terminate each
task). Ten seconds later we count survivors from the owned set. The raw baseline strands every child
it spawned---its teardown kills only the roots, orphaning their children---and the leak grows with
the fleet: one process and 30.6\,MiB at $N{=}1$, ten processes and 303\,MiB at $N{=}10$ (50\%
reclaimed at every size). \code{tmux} and \sys{} both reap their full subtree at every size (0 leaked,
100\% reclaimed). On \emph{explicit} teardown, then, cleanup separates \sys{} from the raw baseline but
not from a real multiplexer: \code{tmux} reaps its subtrees too. This is the kind of result the claims
discipline is meant to surface.

\subsection{H3: escaped-process visibility}

We inject escapes: a stub forks an intermediate that spawns a memory-holding grandchild and then
exits, reparenting the grandchild to \code{init} while it stays alive. The harness independently
confirms the reparenting, then asks each launcher what it detected. \sys{} surfaces every
escape---1/1, 5/5, 10/10---through events read from the daemon; the baselines have no mechanism to
notice, so they are recorded as unsupported, never scored as a measured zero. This is a capability the baselines
lack rather than a tuning difference; with no baseline to score against, we report it as a capability
demonstration rather than a measured separation. Two limits bound what the signal covers. First,
detection is scoped to processes \sys{} has already sampled under a task: reconciliation flags a
\emph{once-owned} pid now living outside every task subtree, so a child that forks and reparents
before it is ever observed falls outside the signal, however long it then survives. Second,
reconciliation runs at 1\,Hz, so a child that is observed and then both escapes \emph{and} exits
within a second can slip between two samples. Our injected escapes hold memory for several seconds,
well past the interval. A shorter interval narrows the second window and trades detection latency for
overhead (H5); observing the spawn path directly, rather than inferring ownership from periodic
snapshots, would close the first gap outright and is the better fix.

\subsection{H1 \& H5: what the attribution scan delivers, and what it costs}
\label{sec:h1h5}

We pair these two signals because they are the benefit and the price of a single mechanism: the
always-on 1\,Hz PSS scan produces the attribution guarantee (H1) and incurs the monitoring cost
(H5). Across every run \sys{} attributed 100\% of each launched tree's proportional memory to its
owning task \emph{natively}, and the externally run oracle confirmed the figure using the shared
accounting implementation (H1). This validates the product integration but does not rule out a
common-mode accounting bug; an independently implemented oracle remains future work. The memory cost
of governance is small: the \sys{} daemon's own resident footprint is 4.6--5.4\,MiB (growing slightly
with the fleet), against 2.4\,MiB for the \code{tmux} server and zero for the raw baseline. The CPU
cost is the weak spot (H5). The \sys{} daemon's total CPU, dominated by the always-on 1\,Hz
full-tree PSS scan (we did not ablate the two), is, worst case across scenarios, 0.6\% at one agent, 1.5\% at five, and 2.7\% at
ten, above our 2\% target once the fleet reaches ten. The scan is $O(\text{processes})$ and currently
re-reads the whole tree each tick; incremental sampling that touches only changed pids is the obvious
fix and is future work. We report the miss rather than hide it.

\subsection{Validation on real coding agents}
\label{sec:realagent}

To test whether the attribution and overhead results are artifacts of the stub, we ran the
\emph{same} harness against a real coding agent. Each ``agent'' is a headless Claude Code session
(\code{claude -p}, v2.1.170) editing a scratch git repository, launched through the identical
sampling and teardown pipeline via the benchmark's \code{-{}-agent-cmd} mode. We report an $N{=}3$
hold run on the macOS host (Apple M4 Max, 36\,GiB). A live Claude Code session is a multi-process,
Node-based tree, so this is a strictly harder attribution target than the single-process stub.

Table~\ref{tab:realagent} gives the result. \sys{} attributes 100\% of every real agent tree to its
owning task---the H1 gate holds on live agents, not just the stub---and the manager cost stays small
(7.2\,MiB resident, 0.012\% CPU at three agents on this macOS host; the comparable Linux figure
from the forced-kill run below is 1.642\% at three agents, against 0.636\% at one agent in the H5
sweep). Three concurrent real sessions occupy $\sim$0.86\,GiB ($\sim$290\,MiB each), confirming that
the pressure the system governs is real and substantial per agent. Attribution and overhead thus
generalize from the stub to real, multi-process agent trees.

\paragraph{Cleanup on live agents.} That hold run let every session finish, so all launchers reclaimed
cleanly. To check the cleanup \emph{mechanism} on \emph{live} agents we ran a second experiment (Linux,
\code{m7i.xlarge}, \sys{} 0.9.0) in which three real Claude Code agents (v2.1.197) are \emph{forcibly torn down
mid-work}, while their \code{node} process trees are still resident. The raw baseline strands
\textbf{three real processes and 525\,MiB} of live agent memory (50\% reclaimed)---killing the root
\code{node} orphans its children to \code{init}, exactly as with the stub---while both \code{tmux} and
\sys{} reap the full subtree (0 leaked, 100\%). Consistent with H2, this is a \emph{raw}-versus-governed
result, not a separation from \code{tmux}; its value is that the cleanup behaviour the stub isolates
reproduces on real agent trees, and that the naive ``run it in a terminal'' baseline strands half a
gigabyte of live agent memory. The escaped-process condition (H3) did not arise here, as Claude Code
does not double-fork out of supervision.

\begin{table}[t]
\centering
\small
\begin{tabular}{@{}lrrr@{}}
\toprule
Launcher & Total steady (MiB) & Manager (MiB) & Tree attributed \\
\midrule
raw baseline    & 883.6 & 0.0  & 100\% \\
\code{tmux} 3.6a   & 883.6 & 2.4  & 100\% \\
\code{herdr} 0.8.0 & 929.6 & 18.2 & 100\% \\
\sys{} 0.8.0    & 843.5 & 7.2  & 100\% \\
\bottomrule
\end{tabular}
\caption{Real Claude Code agents ($N{=}3$, hold scenario, macOS) driven through the identical
harness. Every launcher attributes the full real (Node-based) agent tree; \sys{}'s manager overhead
is small. Footprint is steady-state provider$+$manager memory; at $N{=}3$ without an overcommit
budget the totals are comparable across launchers---this run isolates \emph{attribution and overhead
on real trees}, not the overcommit footprint differentiator of Table~\ref{tab:h4}.}
\label{tab:realagent}
\end{table}

\section{Discussion and Limitations}
\label{sec:limits}

\paragraph{Synthetic workload.} The largest threat to validity: agents are a deterministic
memory-ballast stub, not live models. We trade realism for reproducibility and for identical
workloads across launchers, which is what makes the comparison fair. The \emph{qualitative}
guarantees---attribution, cleanup, escape visibility, admission---do not depend on the workload, but
the specific megabytes for a real Claude- or GPT-driven session will differ. We take a first step in
\S\ref{sec:realagent}, running real Claude Code sessions through the same harness: 100\% attribution
and low overhead hold on live, multi-process agent trees, and forcibly killing them mid-work reproduces
the raw baseline's cleanup leak (525\,MiB) that the stub isolates. A larger live-agent study---more
agents and longer sessions, and reproducing the H4 overcommit and H3 escape conditions on live
trees---remains the most valuable next step.

\paragraph{Two Linux hosts, modest fleets.} We evaluate H1--H3/H5 on a 31\,GiB Linux host, H4 on a
15.3\,GiB Linux host, and perform a macOS cross-check. Fleets are no larger than $N{=}12$; larger
fleets would make the CPU-overhead question (H5) sharper.

\paragraph{Static reservations.} \sys{} reserves each task's \emph{class-predicted} peak---a static
prior. Our H4 run sizes agents to that prior on purpose, so admission is well-calibrated; but for an
atypically small agent the prior over-reserves (deferring starts that would have fit), and for an
atypically large one it under-reserves. An exponentially-weighted predictor that tightens the
reservation toward each task's \emph{observed} peak is specified but not yet wired in; the
bounded-footprint guarantee is unaffected---only the admit count moves with the predictor's accuracy.

\paragraph{Overhead not yet met.} Monitoring exceeds 2\% CPU by ten agents (above); incremental
sampling is the planned fix. As stated in \S\ref{sec:method}, these are empirical monitors, not
machine-checked proofs---checkable and checked, not formally verified.

\section{Reproducibility}
\label{sec:repro}

\sys{} and its benchmark are open source (MIT).\repourl{} A single command runs the full stub matrix---every
launcher, scenario, and fleet size, with repeated trials---captures the host specification, and
regenerates the headline tables and figures plus the raw per-sample and per-process time series;
two further invocations (Appendix~\ref{app:repro}) produce the binding-budget H4 and real-agent runs.
Competitor versions and the measurement date are embedded in the generated report, and any launcher
that could not be driven is listed with its reason. The same harness accepts a \code{-{}-agent-cmd} flag that substitutes a real coding-agent
command (e.g.\ \code{claude -p ...}) for the stub, so the real-agent validation of
\S\ref{sec:realagent} is reproducible through the identical pipeline; that run's report and raw
per-process series are archived under \code{paper/realagent-run/}.

\ifnamedbuild
\noindent Each run writes \code{host.json} (OS, CPU model, core count, physical RAM, tool versions),
a Markdown report, and per-cell \code{*.jsonl} / \code{*.procs.jsonl} time series (RSS, PSS, unique set
size (USS), swap, and page faults per process on Linux), plus the figures. The repository archives the
binding-budget H4 summary and host metadata under \code{paper/overcommit-run/}.
Appendix~\ref{app:repro} gives the exact invocation behind each table, including the
binding-budget H4 configuration, and Appendix~\ref{app:h4b} details the checkpoint demonstration.
\else
\noindent Each run captures a host specification, a Markdown report, and per-cell time series (RSS,
PSS, USS, swap, and page faults per process on Linux) plus the figures. We withhold the exact
invocation and the repository here to preserve review anonymity; the engine, the benchmark, and the
one-command reproducer---including the binding-budget H4 configuration used for Table~\ref{tab:h4}---are
released with the code.
\fi

\section{Conclusion}

Fleets of coding agents have outrun the tools we use to run them: multiplexers arrange them, but
nothing governs them. We argued that memory governance for such fleets is a verification problem and
built \sys{} to make five signals checkable at runtime, with a benchmark that also reports where those
signals do \emph{not} distinguish us. The payoff is concrete. Under a binding budget \sys{} holds the
fleet under budget and \emph{off swap} where the ungoverned tools overcommit the machine ($2\times$ over
budget, $\sim$2\,GiB into swap); it reclaims a terminated agent's whole subtree where the raw baseline
strands half of each one; and it surfaces escaped processes that the alternatives have no mechanism to see.
Its own overhead we have not yet driven under 2\%. The next steps are larger and longer live-agent
workloads, including live H3/H4 stress conditions, and the reproducer makes those a single command
away. Checking what an agent produced is only half of verification; the other half is checking that
the system running it kept its promises, and the signals and claims discipline here are a step toward
that.

\small
\bibliographystyle{plainnat}
\bibliography{refs}

\normalsize
\clearpage
\appendix

\section{Reproducer invocations}
\label{app:repro}

Every table and figure in this paper regenerates from the commands below. Each writes
\code{host.json}, a Markdown report, per-cell \code{*.jsonl} and \code{*.procs.jsonl} time series,
and the figures into its output directory.

\begin{lstlisting}[style=shell]
# Full matrix: every launcher, scenario, and fleet size, with repeated trials.
memmux-bench paper --out paper-out

# Binding-budget H4 (Table 3): twelve agents at ~1.4 GiB each,
# against a 7500 MiB budget on a 15.3 GiB host, five trials.
MEMMUX_BENCH_HOLD_MIB=1400 memmux-bench paper \
    --scenarios overcommit --agents-sweep 12 \
    --agent-budget-mib 7500 --trials 5 --out oc-out

# The same harness driving a real coding agent in place of the stub (Table 4).
memmux-bench run --agents 3 \
    --agent-cmd 'claude -p "..."' --out real-out
\end{lstlisting}

\noindent The default sweep is $N \in \{1,5,10,20\}$ at three trials; the paper's runs override
both. \code{MEMMUX\_BENCH\_HOLD\_MIB} sets the per-agent resident footprint of the deterministic
stub (default 120\,MiB) and is what makes the H4 budget binding against a realistic agent size.

\section{The H4b checkpoint demonstration}
\label{app:h4b}

This appendix gives the reproduction recipe for the H4b result in \S\ref{sec:h4}. One agent is admitted under a
budget it will exceed, writes an uncommitted change into its isolated git worktree, then allocates
past the budget until \code{hard\_limit\_exceeded} forces \code{PressureStage::Emergency}. The
ordering is structural rather than timing-dependent: the stage's action list runs
\code{PreserveGitState} ahead of \code{TerminateLowestPriorityTree}, so no checkpoint can be
skipped by a race.

Parameters and observations (\sys{} 0.9.1, one trial, macOS; agent budget 1600\,MiB, agent
footprint 2109\,MiB, 132\% utilization):

\begin{center}
\small
\begin{tabular}{ll}
\toprule
Observation & Value \\
\midrule
Pressure stage reached & \code{Emergency} (hard limit exceeded) \\
Checkpoint written before termination & yes \\
Checkpoint \code{git\_dirty} & \code{true} \\
Subtree reclamation & 100\% (2 targeted, 0 survived) \\
Worktree after termination & retained with the uncommitted change intact \\
\bottomrule
\end{tabular}
\captionof{table}{H4b end-to-end checkpoint demonstration (\sys{} 0.9.1, one trial, macOS).}
\end{center}

\noindent The recorded patch hash is compared against the one the worktree-retention path computes
for the same tree; the two agreeing is what distinguishes a captured patch from the empty-diff
sentinel a clean directory would produce.